%% file: GFTNN_v2.tex
\documentclass[letterpaper, 10 pt, conference]{ieeeconf}  

\IEEEoverridecommandlockouts                              

\usepackage{amsmath} 
\usepackage{amssymb}  

\usepackage{graphicx,import}

\usepackage{cite}
\usepackage{algorithmic}
\usepackage{graphicx}
\usepackage{textcomp}
\usepackage{xcolor}
\usepackage{bm}
\usepackage{multirow}
\usepackage{siunitx}
\usepackage{tikz}
\usepackage{pgfplots}
\usepackage{subcaption}
\graphicspath{{./fig/}}
\usetikzlibrary{arrows}
\usetikzlibrary{shapes}
\usetikzlibrary{arrows, calc,intersections}
\usetikzlibrary{arrows.meta}
\usetikzlibrary{decorations.pathmorphing}
\usetikzlibrary{decorations.text}
\usetikzlibrary{decorations.pathreplacing}
\usetikzlibrary{hobby,decorations.markings}
\usetikzlibrary{positioning}
\usetikzlibrary{calc}
\usetikzlibrary{fadings}
\tikzset{%
	every neuron/.style={
		circle,
		draw,
		minimum size=0.15cm
	},
	neuron missing/.style={
		draw=none, 
		scale=1,
		text height=1,
		execute at begin node=\color{black}$\vdots$
	},
}

\usepackage{pifont}
\usepackage[switch]{lineno}  
\usepackage[hyphens]{url}  

\def\BibTeX{{\rm B\kern-.05em{\sc i\kern-.025em b}\kern-.08em
		T\kern-.1667em\lower.7ex\hbox{E}\kern-.125emX}}

\usepackage{textcomp}
\newcommand\copyrighttext{%
	\footnotesize \textcopyright 2023 IEEE. Personal use of this material is permitted. Permission from IEEE must be obtained for all other uses, in any current or future media, including reprinting/republishing this material for advertising or promotional purposes, creating new collective works, for resale or redistribution to servers or lists, or reuse of any copyrighted component of this work in other works. DOI: t.\,b.\,d.}
\newcommand\copyrightnotice{%
	\begin{tikzpicture}[remember picture,overlay]
	\node[anchor=south,yshift=10pt] at (current page.south) {\fbox{\parbox{\dimexpr\textwidth-\fboxsep-\fboxrule\relax}{\copyrighttext}}};
	\end{tikzpicture}%
}
\begin{document}

\title{\LARGE \bf
Prediction and Interpretation of Vehicle Trajectories\\in the Graph Spectral Domain}
\author{
	Marion Neumeier\textsuperscript{\rm 1}*\thanks{*We appreciate the funding of this work by AUDI AG.},
	Sebastian Dorn\textsuperscript{\rm 2},
	Michael Botsch\textsuperscript{\rm 1} and 
	Wolfgang Utschick\textsuperscript{\rm 3}
	\thanks{\textsuperscript{\rm 1} CARISSMA Institute of Automated Driving, Technische Hochschule Ingolstadt, 85049 Ingolstadt, Germany {\tt\small firstname.lastname@thi.de}}%
	\thanks{\textsuperscript{\rm 2} AUDI AG, 85057 Ingolstadt, Germany {\tt\small sebastian.dorn@audi.de}}	
	\thanks{\textsuperscript{\rm 3} Technical University of Munich, 80333 Munich, Germany {\tt\small utschick@tum.de}}
}
\maketitle
\copyrightnotice
\begin{abstract}
This work provides a comprehensive analysis and interpretation of the graph spectral representation of traffic scenarios. Based on a spatio-temporal vehicle interaction graph, an observed traffic scenario can be transformed into the graph spectral domain by means of the multidimensional Graph Fourier Transformation. Since these spectral scenario representations have shown to successfully incorporate the complex and interactive nature of traffic scenarios, the beneficial feature representation is employed for the purpose of predicting vehicle trajectories. This work introduces GFTNNv2, a deep learning network predicting vehicle trajectories in the graph spectral domain. Evaluation of the GFTNNv2 on the publicly available datasets highD and NGSIM shows a performance gain of up to $\mathbf{25\,\%}$ in comparison to state-of-the-art prediction approaches.
\end{abstract}

\section{Introduction}
Motion forecasting of traffic participants has been gaining increasing attention, as it is a crucial task in the development of autonomous driving systems. Based on the observed states of traffic participants, and potentially road map information, the goal is to predict the future trajectory of an agent or all involved traffic participants. However, trajectory prediction can be extremely challenging since traffic scenarios are particularly influenced by interdependencies between the individual participants. These crucial interdependencies arise from temporal and spatial relations within a traffic scenario. Since explicitly modeling these bidirectional relations and their effects is challenging, state-of-the-art methods for trajectory prediction are predominantly based on Machine Learning (ML) \cite{Deo.2018, Alahi.2016, Cao.5302021652021, Neumeier.2022}. The recent success of Graph Neural Networks (GNNs) in various domains has also boosted research on their applicability in the automotive domain and trajectory prediction, in particular\cite{Gao.08.05.2020, RAGAT.2021}. GNNs are ML networks that operate on graph structures. As interdependencies between traffic participants can be modeled by edges in graph structures \cite{Gao.08.05.2020, S.Suo.2021}, GNNs enable to explicitly incorporate such relational information. 
Similar to GNNs, the Graph Fourier Transformation Neural Network (GFTNN) proposed in \cite{Neumeier.2022} is a statistical learning model leveraging connectivity information of graph structures. The GFTNN transforms a spatio-temporal traffic scenario graph into a multidimensional graph spectral scenario representation and uses this representation as input for an end-to-end learnable prediction module. Evaluation in the task of highway trajectory prediction revealed the promising performance of the GFTNN.
In the work of \cite{Neumeier.2022}, the authors provide insight on the properties and interpretation of the spectral scenario representation. However, while their work thoroughly elaborates the interpretation of the temporal dimension within the spectral scenario representation, the spectrum's spatial dimension was out of scope. 

In this work, the goal is to address this open research question and analyze the interpretation of the spatial dimension within the multidimensional graph spectral scenario representation. Interpreting and understanding feature representations is relevant because it can provide insights into underlying patterns and relationships in complex data. Such insights can be leveraged to introduce sophisticated network architectures or loss terms that improve performance \cite{Snderby.2016, Revach.2022}.
In addition, this work introduces GFTNNv2 which is an enhanced version of GFTNN\cite{Neumeier.2022} that shows competitive performance against its predecessor and state-of-the-art models in the task of highway trajectory prediction. In comparison to GFTNN, GFTNNv2 is designed in such a fashion that it does not have to (implicitly) learn the inverse multidimensional Graph Fourier Transform (GFT). The new architecture also allows for more complex predictions, as the motion paths of GFTNN were restricted in flexibility due to the descriptive decoder. 
%
%


\textbf{Contribution.} This work contributes towards a better understanding of the graph spectral representation of traffic scenarios using the multidimensional GFT and its usage to improve performance in long-term trajectory predictions.\\
The main contributions are as follows:
\begin{itemize}
	\item Analysis and interpretation of the graph spectral scenario representation generated through the multidimensional GFT with a focus on the spatial dimension.
	\item Introduction of GFTNNv2, a vehicle trajectory prediction network for highway scenarios.
	\item Performance evaluation of GFTNNv2 and comparison to state-of-the-art prediction models on the publicly available highway datasets highD and NGSIM. 
\end{itemize}
In this work, vectors are denoted as bold lowercase letters and matrices as bold capital letters.
\section{Related Work}
Vehicle trajectory prediction is a field of research with a significant number of works aimed at improving the accuracy and robustness of predictions. This section focuses on reviewing ML-based prediction networks, but there are various other approaches that utilize model-based methodologies to predict maneuvers \cite{Elter.2022, SimonUlbrich.2013, Framing.2018}. For a comprehensive survey, refer to the works authored by Lefèvre \textit{et al.} \cite{Lefevre.2014} or Manish \textit{et al.} \cite{Manish.2023}.

Due to the sequential nature of trajectory prediction tasks, the majority of deep learning prediction networks are based on Recurrent Neural Networks (RNNs) \cite{STAM-LSTM, Alahi.2016, Deo.2018, Neumeier.2021, Messaoud.2021, NLS-LSTM}. Well-established methods like Social  Long Short-Term Memory (Social LSTM) \cite{Alahi.2016} and Convolutional Social Pooling LSTM (CS-LSTM) \cite{Deo.2018}, for example, use RNNs and Convolution Neural Networks (CNNs) to evaluate the sequential and interdependent nature of the trajectory prediction task. Since RNNs lack to capture social interactions, these methods include social pooling strategies to decode interactions. In \cite{Messaoud.2021}, the authors introduced the Multi-head Attention Social Pooling (MHA-LSTM) model, which applies a multi-head attention mechanism \cite{Vaswani.2017} to relate also distant traffic participants. A prediction network utilizing a sequence-to-sequence architecture and a Generative Adversarial Network (GAN) is proposed in \cite{Gupta.2018}. While performing well, recurrent methods suffer from time-consuming iterative propagation and gradient explosion or vanishing issues \cite{Pascanu.21.11.2012}. 

Recently, statistical learning architectures based on graph structures have grown in popularity for predicting future trajectories \cite{STAM-LSTM, Mo.08.07.2021, Neumeier.2022, RAGAT.2021, MALS2023, Cao.5302021652021, Gao.08.05.2020}. These so-called GNNs are used to directly incorporate temporal and spatial interactions of traffic participants as well as map features. GNN approaches mainly differ by the definitions of the used aggregation function and update function\cite{Zhou.2020, Wu.2021}. The authors of \cite{Yu.2018} proposed the Spatio-Temporal Graph Convolution Neural Network (STGCN) to tackle the task of motion prediction, where observed traffic scenarios are represented through a stacked set of spatial graphs. To fuse features from both domains, a spatio-temporal convolutional block is presented. Cao \textit{et al.} \cite{Cao.5302021652021} introduced the Spectral Temporal Graph Neural Network (SpecTGNN), which separately models the environment and vehicle interactions using graph structures. By successively applying a spectral graph convolution and a temporal convolution, motion information of the interactive agents (and the environment) is encoded. In \cite{Zhou.2021}, the authors introduced the Attention-based Spatio-Temporal Graph Neural Network (AST-GNN) for pedestrian trajectory prediction. To handle the multidimensionality of the task, AST-GNN is composed of a spatial GNN and temporal GNN. Additionally, the attention mechanism is used to identify interactions between pedestrians. In general, attention-based networks and Graph Attention Networks (GATs) have been gaining traction recently. The authors of \cite{RAGAT.2021}, for example, introduced the Repulsion and Attraction Graph Attention (RA-GAT) for vehicle trajectory prediction. RA-GAT follows the idea of repulsive and attractive forces within a traffic scenario, which are modeled using GATs\cite{GAT.2018}. To encode and decode the underlying vehicle movements, LSTMs are used. Wu \textit{et al.} \cite{HSTA.2021} suggested a Hierarchical Spatio-Temporal Attention (HSTA) model for tackling the task of trajectory prediction. The proposed network combines GATs, sequential MHAs and LSTMs to capture spatio-temporal interactions and predict trajectories. However, as shown in \cite{Neumeier.2023, Neumeier.arxiv2023}, commonly used GAT architectures \cite{GAT.2018, GATv2} are sensitive to network initialization. Sparse and small graphs, such as those in automotive settings, can experience early stagnation in parameter training.

Contrary to many recently suggested prediction networks, the GFTNN \cite{Neumeier.2022} does not rely on sophisticated or recurrent learning architectures but a lean Feedforward Neural Network (FNN). GFTNN is a lightweight prediction model, that transforms spatio-temporal traffic scenarios into the graph spectral domain using the multidimensional GFT. The spectral traffic scenario representation is used as input for the trajectory prediction of the encoder-decoder based architecture. The FNN within the encoder learns to map the spectral traffic scenario representation into a latent representation, which is used by the descriptive decoder \cite{Neumeier.2021} to perform the trajectory prediction.

This work follows on from the efforts of \cite{Neumeier.2022}. In the subsequent sections, the graph spectral traffic scenario representation is interpreted and network's prediction architecture is refined.
\section{Preliminaries}
\textbf{Graph definition}: 
Let $\mathcal{G} = (\mathcal{V},\mathcal{E})$ be a graph composed of $N=|\mathcal{V}|$ nodes and $E = |\mathcal{E}|$ edges. The connectivity of a graph is typically represented by the adjacency matrix \mbox{$\mathbf{A} \in \mathbb{R}^{N\times N}$}. The matrix element  $\mathbf{A}_{ij} = 1$ if an edge connects node $i$ to $j$, and $\mathbf{A}_{ij} = 0$ otherwise. In this work, each graph is undirected, and hence, its corresponding adjacency matrix is symmetric $\mathbf{A}_{ij} = \mathbf{A}_{ji}$. The multidimensional GFT~\cite{Kurokawa.2017} is based on the graph representation by the Laplacian matrix $\mathbf{L} = \mathbf{D} - \mathbf{A}$, where $\mathbf{D} \in \mathbb{R}^{N\times N}$ is the degree matrix. The degree matrix is a diagonal matrix with the node degrees on the diagonals such that $\mathbf{D}_{ii} = \sum_{j} \mathbf{A}_{ij}$.\\

\begin{figure}
	\centering
	\vspace{9pt}
	\def\svgwidth{0.85\columnwidth}
	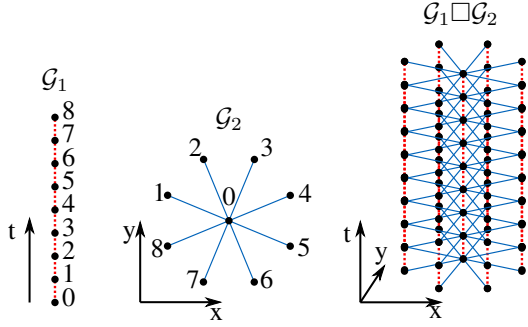
	\caption{Cartesian product graph $\mathcal{G}_1\square \mathcal{G}_2$ of the factor graphs $\mathcal{G}_1$ and $\mathcal{G}_2$. Red dotted edges in $\mathcal{G}_1 \square  \mathcal{G}_2$ result from graph $\mathcal{G}_1$ and blue solid edges come from $\mathcal{G}_2$ \cite{Neumeier.2022}.}
	\label{fig:CartesianProductGraph}
	\vspace{-10pt}
\end{figure}

\textbf{Multidimensional graph Fourier transformation}:
The multidimensional GFT transforms signals on Cartesian product graphs into a multidimensional frequency domain. In graph theory, a Cartesian product graph $\mathcal{G}_1\square \mathcal{G}_2$ refers to the product operation of the factor graphs $\mathcal{G}_1 = (\mathcal{V}_1,\mathcal{E}_1)$ and $\mathcal{G}_2= (\mathcal{V}_2,\mathcal{E}_2)$ as qualitatively illustrated in Fig.~\ref{fig:CartesianProductGraph}. 
The multidimensional GFT $\mathcal{F}(\mathcal{G}_1 \square \mathcal{G}_2, \mathbf{F})$ of a Cartesian product graph $\mathcal{G}_1 \square \mathcal{G}_2$ with the signal $\mathbf{F}: \mathcal{V}_1 \times  \mathcal{V}_2 \rightarrow \mathbb{R}$ is defined as 
\begin{equation}
\mathcal{F}(\mathcal{G}_1 \square \mathcal{G}_2, \mathbf{F}) = \mathbf{\hat{F}} = 
\mathbf{U}_1^*
\mathbf{F}
\overline{\mathbf{U}}_2,
\label{eq:GFTmat}
\end{equation}
and its inverse is given by
\begin{equation}
\hat{\mathcal{F}}(\mathcal{G}_1 \square \mathcal{G}_2, \mathbf{F})=
\mathbf{{F}} = 
\mathbf{U}_1
\mathbf{\hat{F}}
{\mathbf{U}}_2^\text{T},
\label{eq:invGFTmat}
\end{equation}
where $\mathbf{U}_n \in \mathbb{R}^{N_n \times N_n}$ with $N_n = |\mathcal{V}_n|$ is the eigenbasis (set of eigenvectors) arising from each factor graph \mbox{$n = 1,2$\cite{Kurokawa.2017}}. $\mathbf{U}_n^*$ denotes the Hermitian transpose and $\overline{\mathbf{U}}_n$ is the element-wise complex conjugate matrix. The graph signal $\mathbf{F}\in \mathbb{R}^{N_1\times N_2}$ holds the feature information for each node in $\mathcal{G}_1 \square \mathcal{G}_2$.
The eigenbases are computed through the eigendecomposition of the Laplacian matrices $\mathbf{L}_n$ such that
\begin{equation}
\mathbf{L}_n = \mathbf{U}_n\mathbf{\Lambda}_n\mathbf{U}_n^{-1},
\label{eq:EIG}
\end{equation}
where the $i$-th column of the square matrix $\mathbf{U}_n$ is the eigenvector $\bm{u}_i^{(n)}\in \mathbb{R}^{N_n}$ of $\mathbf{L}_n$, and $\mathbf{\Lambda}_n$ is a diagonal matrix with the eigenvalues on its diagonal $\mathbf{\Lambda}_{n, ii}= \lambda_i^{(n)}$. Since the Laplacian matrix is real, symmetric and positive-semidefinite, its eigendecomposition results in $N_n$ real, non-negative eigenvalues $0=\lambda_0^{(n)},\dots, \lambda_{N_n-1}^{(n)}=\lambda_\mathrm{max}^{(n)}$. The resulting set of eigenvalues is usually sorted by magnitude. 
The forward and inverse GFT depend on the choice of eigenvectors, which does not necessarily have a unique solution~\cite{DavidIShuman.2016}. Throughout this work, the choice is not particularized but the determination of eigenvectors is assumed to be fixed.

\begin{figure*}[ht]
	\vspace{4pt}
	\centering
	\input{tikz/pred_model}
	\vspace{-1pt}
	\caption{In GFTNNv2, the spatio-temporal graph $\mathcal{G}_\textrm{S}\square\mathcal{G}_\textrm{T}$ is used to represent interdependencies within the observed traffic scenario $\mathbf{F}$. By applying the GFT, the traffic scenario $\mathbf{F}$ is transformed into a graph spectral representation $\mathbf{\hat{F}}_\textrm{obs}$. The prediction module uses this spectral representation $\mathbf{\hat{F}}_\textrm{pred}$ to predict the scenario evolution in the graph spectral domain using the FNN $g_\theta:\mathbb{R}^{Z} \rightarrow \mathbb{R}^{2\times N\times q}$. Subsequently, $\mathbf{\hat{F}}_\textrm{pred}$ is converted to the spatio-temporal domain by applying the inverse GFT.}
	\label{fig:gilbert_architecture}
	\vspace{-8pt}
\end{figure*}
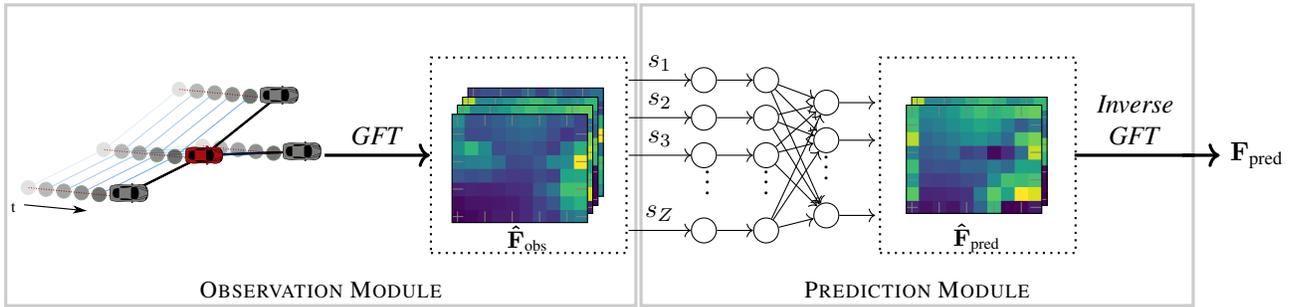

\section{Method}
This section initially introduces the general concept of the GFTNNv2, followed by a detailed discussion of the model's architecture.
In Fig.~\ref{fig:gilbert_architecture}, the overall architecture of the GFTNNv2 is shown. The network is structured into two key components, namely the observation module and the prediction module. Similar to GFTNN\cite{Neumeier.2022}, the observation module of GFTNNv2 computes the spectral scenario representation based on the observed traffic scenario. In GFTNN, the spectrum is used to predict the trajectory of the target vehicle in the spatio-temporal domain directly. Since this prediction inherently includes a transformation from the graph spectral domain to the spatio-temporal domain, the applied prediction network has to learn the mapping between the domains. However, as the mathematical function to perform this mapping is well defined by the inverse GFT, the learning task for the network is unnecessarily complex. The idea of the GFTNNv2 is to avoid the additional learning complexity: Instead of predicting the spatio-temporal features of the scenario, GFTNNv2's prediction module predicts the future scenario in the graph spectral domain. By applying the inverse GFT on the predicted spectrum, the regular spatio-temporal scenario representation is generated. The resulting representation holds the predicted trajectory for all observed traffic participants. Hence, rather than predicting the trajectory of one target agent as in GFTNN\cite{Neumeier.2022}, GFTNNv2 implicitly predicts the complete scenario evolution w.\,r.\,t. all considered traffic participants. However, to allow a fair comparison between both approaches, in this work performance evaluation is based on single agent prediction. Therefore, the implementation details of GFTNNv2 are designed for making predictions for a single agent, as described in the following.
\subsection{Observation Module}
The observation module aims to convert a traffic scenario into a feature representation in the graph spectral domain. Therefore, the observed traffic scenario is represented as a Cartesian product graph based on the temporal and spatial interdependencies as proposed in \cite{Neumeier.2022}. By means of this interaction graph, the scenario is transformed into a spectral representation using the multidimensional GFT.

The Cartesian product graph $\mathcal{G}_\text{S} \square  \mathcal{G}_\text{T}$ is composed of a spatial factor graph $\mathcal{G}_\text{S}$ and a temporal factor graph $\mathcal{G}_\text{T}$, which describe the interaction assumptions within each dimension. The factor graph $\mathcal{G}_\text{S} = (\mathcal{V}_\text{S}, \mathcal{E}_\text{S})$ represents the spatial relations between the $N = |\mathcal{V}_\text{S}|$ traffic participants. Since the immediately adjacent traffic participants of a target agent are considered the most relevant for the trajectory prediction, $\mathcal{G}_\text{S}$ is defined as star graph \cite{Neumeier.2022}. Hence, all traffic participants primarily interact with the target vehicle, but not among each other. While there also might be interdependencies between neighboring agents these are not explicitly modeled. The spatial graph definition is similar to graph $\mathcal{G}_\text{2}$ in Fig.~\ref{fig:CartesianProductGraph}. Furthermore, as each vehicle has dependencies with itself as time evolves, the temporal factor graph $\mathcal{G}_\text{T}= (\mathcal{V}_\text{T}, \mathcal{E}_\text{T})$ is defined as a linear graph, where $H = |\mathcal{V}_\text{T}|$ is the total number of temporal observations. The value of $H=T_\textrm{obs}\cdot f$ depends on the sample rate $f$ and the observed time span~$T_\textrm{obs}$. Each time step of the observation is represented as a node, which is connected to its previous and following node. The temporal factor graph $\mathcal{G}_\text{T}$ is qualitatively similar to graph  $\mathcal{G}_\text{1}$ in Fig.~\ref{fig:CartesianProductGraph}. Based on the introduced interaction assumptions, the Cartesian product graph is defined by the temporal Laplacian matrix \mbox{$\mathbf{L}_\text{T} \in \mathbb{R}^{H \times H}$} and spatial Laplacian matrix \mbox{$\mathbf{L}_\text{S} \in \mathbb{R}^{N\times N}$}.

For each time step, a traffic participant holds $K$ features describing its state, e.\,g., its position or velocity. Hence, the graph signal describing the observed traffic scenario is defined as $\mathbf{F} \in \mathbb{R}^{K \times N \times H}$. For each feature $k = 1, \dots, K$, the multidimensional GFT is computed through
\begin{equation}
	\mathbf{\hat{F}}_{\text{obs}, k}= \mathbf{U}_\text{S}^*
	\mathbf{F}_k
	\overline{\mathbf{U}}_\text{T},
	\label{eq:implemented}
\end{equation}
where $\mathbf{U}_\text{S} \in \mathbb{R}^{N \times N}$, $\mathbf{U}_\text{T} \in \mathbb{R}^{H \times H}$ are the resulting eigenbases based on Eq.~\ref{eq:EIG} and $\mathbf{\hat{F}}_{\text{obs}, k}\in \mathbb{R}^{N\times H}$ is the spectral scenario representation of the feature $k$. 

Since in GFTNNv2 the bases $\mathbf{U}_\text{S}, \mathbf{U}_\text{T}$ are not only used for the forward GFT but also for the inverse GFT in the prediction module, a potential dimensional inconsistency must be addressed. If the observation period and prediction period differ $T_\textrm{obs} \neq T_\textrm{pred}$, the inverse GFT cannot be computed due to deviating temporal feature dimensions \mbox{$H \neq H_\textrm{pred}$}, where $H_\textrm{pred}=T_\textrm{pred}\cdot f$ defines the number of total prediction steps. To prevent this, the features $\mathbf{F}$ obtained from the observed traffic scenario are upsampled to \mbox{$H=H_\textrm{pred}$} before constructing the graph (as usually the prediction period is longer than the observation period) and the temporal factor graph $\mathcal{G}_\text{T}$ is adapted analogously. 
Similar to the conventional Fourier transform, the resulting graph spectrum $\mathbf{\hat{F}}_{\text{obs}, k}$ characterizes the signal frequencies for feature $k$ on the graph: Low eigenvalues are associated with lower frequencies, indicating that feature differences between neighboring nodes are small \cite{Neumeier.2022}. Note however that the eigenvalues are not necessarily equivalent to the physical definition of the temporal frequency. As shown in \cite{Neumeier.2022}, information of large eigenvalues in the temporal domain of the spectrum can (potentially) be neglected without loss of decisive information. By considering only the information of the $p \leq H$ lowest temporal eigenvalues $\lambda_0^{(T)}, \dots, \lambda_{p-1}^{(T)} $ in the graph spectral domain $\mathbf{\hat{F}}_{\text{obs}, k}$, the inverse transformation results in a low-pass filtered version of feature $k$ in the original time domain. Since in the application of trajectory prediction the relevant information lies in the small temporal eigenvalues, neglecting high frequency information does not crucially impede prediction performance, but reduces the learning task complexity. Hence, if computational resources are limited, setting $p<H$ helps to reduce the number of learnable parameters and optimize run-time. As the effect of temporal spectral filtering in the graph spectral domain has already been addressed extensively in \cite{Neumeier.2022}, an ablation study on the hyper-parameter of $p$ is not a focus of this work. The parameter $p$ is set to $H$ and is not altered during evaluation. 
Generally, however, in GFTNNv2 a selected subset \mbox{$\mathcal{S}_p \subseteq \mathbf{\hat{F}}_{\text{obs}}$} depending on the hyper-parameter $p$ is passed to the prediction module as input. The higher the value of $p$, the more information about the temporal frequencies is considered. The selected subset $\mathcal{S}_p \in \mathbb{R}^{K\times N \times p}$ is reshaped to a vector representation $\bm{s}_p \in \mathbb{R}^Z$, where $\bm{s}_p = [s_1, \dots, s_Z]^\text{T}$ and $Z= |K \times N \times p|$. 

\subsection{Prediction Module}
Based on the subset $\mathcal{S}_p$, the prediction module predicts the spectral representation $\mathbf{\hat{F}}_{\text{pred}}\in\mathbb{R}^{2\times N \times H_\textrm{pred}}$ of the future scenario. $\mathbf{\hat{F}}_{\text{pred}}$ characterizes the future behavior of the $N$ considered traffic participants in the graph spectral domain. Similar to the hyper-parameter $p$ in the observation module, in GFTNNv2 a hyper-parameter $q$ is introduced to limit the number of predicted frequencies $q\leq H_\textrm{pred}$. By setting \mbox{$q< H_\textrm{pred}$}, only the amplitudes of the $q$ lowest temporal frequencies are predicted while the remaining frequency amplitudes are set to zero. In other words, the parameter $q$ defines the dimension of the predicted subset $\mathbf{\hat{F}}_{\text{q}}\subseteq \mathbf{\hat{F}}_{\text{pred}}$, where $\mathbf{\hat{F}}_{\text{q}}\in\mathbb{R}^{2\times N \times q}$. The mapping $g_\theta: \mathcal{S}_p \rightarrow \mathbf{\hat{F}}_{\text{q}}$ is performed by a FNN, which does not contain computationally expensive operations. The prediction network is designed as plain FNN composed of three layers ($Z - |K\times N\times 32| - |2\times N\times q|$). The GELU activation function is used between the first two layers, and the sigmoid activation function is used between the second and last layer.
By applying the inverse multidimensional GFT to the predicted spectral scenario $\mathbf{\hat{F}}_{\text{pred}}$ using
\begin{equation}
	\mathbf{F}_{\text{pred}, l}  = 
	\mathbf{U}_\text{S}
	\mathbf{\hat{F}}_{\text{pred}, l}
	{\mathbf{U}}_\text{T}^\text{T},
\end{equation}
where $l=1,2$, the scenario representation is transformed from the graph spectral domain back into the spatio-temporal domain. Hence, $\mathbf{F}_{\text{pred}} \in \mathbb{R}^{2\times N \times H_\textrm{pred}}$ holds the longitudinal and lateral trajectory information of the $N$ traffic participants for the upcoming $H_\textrm{pred}$ prediction steps. The trajectory prediction $\mathbf{Y}_\textrm{pred}^{(i)} \in \mathbb{R}^{2\times H_\textrm{pred}}$ of traffic participant \mbox{$i \in \{1, \dots, N\}$} is determined by selecting the $i$-th entry of $\mathbf{F}_{\textrm{pred}}^{(i)}$.

In comparison to GFTNN \cite{Neumeier.2022}, the architecture of this work has the advantage, that the FNN does not have to learn the mapping from the graph spectral domain to the spatio-temporal domain. Thereby, complexity of the learning task is reduced. Furthermore, GFTNNv2 inherently provides the predicted trajectory of all $N$ considered traffic participants instead of performing single agent trajectory prediction only. 

\section{Interpretation of the Graph Spectral Domain}
\begin{figure*}[t!]
	\centering
	\includegraphics[width=\linewidth]{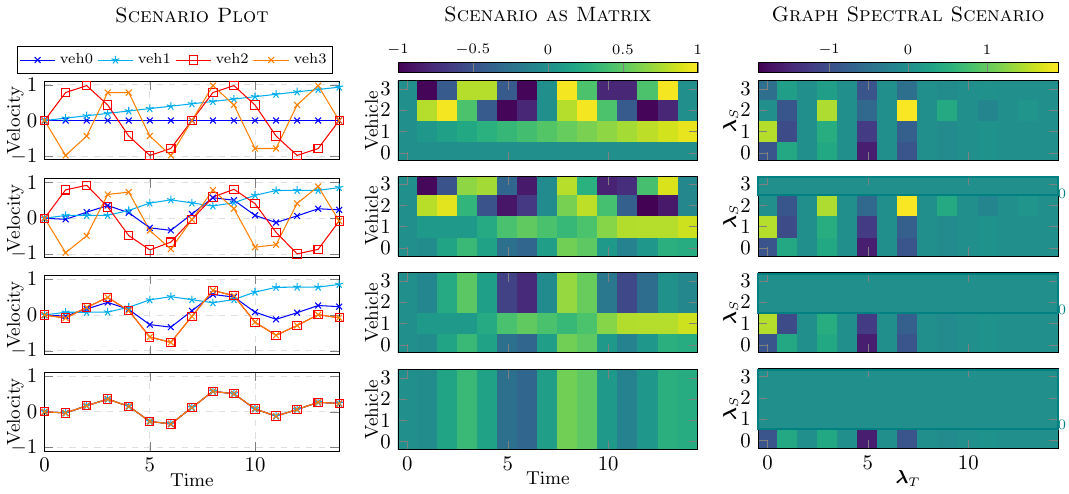}
	\vspace{-12pt}
	\caption{Effect of spatial frequency filtering in the graph spectral domain: By neglecting an increasing number of spatial frequencies $\lambda_{S}$ (top to bottom) in the graph spectral scenario (right column), the resulting scenario (left column) progressively varies from the initial scenario. The central column shows the filtered scenario from the left column as matrix representation.}
	\label{fig:spectralFiltering}
	\vspace{-11pt}
\end{figure*}

\input{tikz/spectralfiltering.tex}
The multidimensional GFT transforms graph signals into spectral representations while containing the multidimensional context. These spectral representations indicate the quantity of how much a graph signal varies along each graph dimension or, in other words, how smooth the graph signal is. Thus, when representing a traffic scenario through a graph structure and applying the multidimensional GFT, the resulting spectral representation quantifies the dynamics of a vehicle feature within the temporal and spatial domain. An exemplary graph spectral scenario representation for a scalar feature can be seen in Fig.~\ref{fig:spectralFiltering}: The uppermost left plot illustrates the velocity information of four different vehicles in a synthetic traffic scenario. By representing the scenario using the proposed Cartesian Product $\mathcal{G}_\text{S} \square  \mathcal{G}_\text{T}$ with \textit{veh0} being the central node and applying the multidimensional GFT, the graph spectral scenario representation shown in the uppermost right plot is generated. The spectral representation characterizes the scenario as composition of spatial frequencies ($\bm{\lambda}_S$) and temporal frequencies ($\bm{\lambda}_T$). However, while the general intuition about the spectrum can be carried over from the sequential Fourier transform, the frequencies in the graph setting do not necessarily indicate sinusoidal basis functions. Only under certain conditions, the common physical notion of frequency is preserved. The eigenvalues (or frequencies) and eigenvectors used as basis functions in the GFT depend on the graph connectivity. In the case of an unweighted path graph, the graph Laplacian eigenvectors are identical to the basis vectors in the Discrete Cosine Transform \cite{Strang.1999, DavidIShuman.2016}. Since the temporal factor graph $\mathcal{G}_\text{T}$ is an unweighted path graph, the interpretation of the temporal frequencies and applied filtering operations within the spectral representation are highly intuitive\cite{Neumeier.2022}. On the contrary, the interpretation of the spatial frequencies in the graph spectral domain is less unambiguous. The spatial factor graph $\mathcal{G}_\text{S}$ is defined as star graph. As can be seen in Fig.~\ref{fig:sinusoidalApprox}, the Laplacian eigenvectors of a star graph with four nodes do not form a sinusoidal basis. The values of the eigenvectors can only partially be approximated by a sinusoidal function. Yet, eigenvectors associated with high eigenvalues indicate more zero crossings and consequently correlate with the general intuition of frequency.
Setting the values associated with an eigenvector $\bm{\lambda}_{S,i}$ in the graph spectral representation of a traffic scenario to zero equals frequency filtering in the spatial domain. In contrast to filtering operations in the temporal domain\cite{Neumeier.2022}, filtering of spatial frequencies alters the feature differences of the vehicles at each given time step w.\,r.\,t. each other. In other words, filtering out the highest spatial frequency results in a reduction of the largest feature differences over the complete time span in the original scenario. For example, in the second row of Fig.~\ref{fig:spectralFiltering}, the highest spatial frequency $\bm{\lambda}_{S,i=3}$ of the synthetic traffic scenario is filtered out by setting the associated values in the graph spectral domain to zero. In the original scenario domain (left column) this filtering particularly causes the velocity of \textit{veh0} and \textit{veh1} to be oscillating over time. Although this might be counter-intuitive first, by analyzing the effect based on the scenario as matrix plot in Fig.~\ref{fig:spectralFiltering} (central column) this outcome becomes more clear. The spatial frequencies describe the feature variation among the vehicles at each time step. Hence, high feature variation along the vertical axis (columns) of the scenario matrix plot are associated with high frequencies. Consequently, filtering out high frequencies in the spatial domain causes the vehicles' velocities to become increasingly similar at each time step and mitigate feature differences. Through gradually increasing the number of filtered frequencies in the spatial domain, an increasing alignment or similarity in the features of the individual vehicles is observable. As can be seen in the bottom row of Fig.~\ref{fig:spectralFiltering}, when filtering all spatial frequencies except $\bm{\lambda}_{S,i=0}$ the velocity of all vehicles are similar for each time step. While there is still a feature oscillation over time, no feature oscillation remains in the vertical axis of the scenario as matrix plot. Each vehicle has the same value for each time step or column. 
From this analysis it can be concluded, that while filtering in the temporal domain is useful for automotive trajectory prediction tasks\cite{Neumeier.2022}, filtering in the spatial domain is disadvantageous for most of the use cases since relevant information is corrupted or lost. Hence, while the proposed GFTNNv2 provides a hyper-parameter $q$ for filtering certain temporal frequencies, the introduction of a similar parameter for the spatial domain is not expedient.

\section{Experimental Setup and Evaluation}
Although GFTNNv2 is able to perform multi-agent predictions, the evaluation is based on single agent predictions to allow a fair comparison with the previously introduced GFTNN. GFTNN is designed to perform only single agent predictions. Additionally, the proposed model is benchmarked against state-of-the-art prediction models.
\subsection{Problem Definition}
Based on the motion observation of $N=9$ vehicles for the time span $T_\textrm{obs}=\SI{3}{\second}$, the task is to predict the trajectory of the central target vehicle $i \in \{1, \dots, N\}$. The prediction horizon for the trajectory is set to $T_\textrm{pred}=\SI{5}{\second}$. In total, $K=4$ vehicle features are considered during the observation period such that the motion information of the \mbox{$j$-th} participating road user is ${\boldsymbol{\xi}_j= [\bm{x}_{j}, \bm{y}_{j}, \bm{v}_{j,\textrm{x}}, \bm{v}_{j,\textrm{y}}]^\textrm{T}}$ which contains the past longitudinal and lateral positions ($\bm{x}_{j}, \bm{y}_{j}$) and velocities ($\bm{v}_{j,\textrm{x}}, \bm{v}_{j,\textrm{y}}$) up to the current time step $ t_0 $. Depending on the dataset's sample rate $f$, the model input is ${\mathbf{X} = [\boldsymbol{\xi}_1, \boldsymbol{\xi}_2, \dots , \boldsymbol{\xi}_{N}]^\textrm{T}}$, where $\boldsymbol{\xi}_j \in \mathbb{R}^{K \times H}$ and $H=T_\textrm{obs}\cdot f$. Based on the observed traffic scenario $\mathbf{X}$, the network predicts the trajectory of the target vehicle \mbox{${\mathbf{Y}}_\textrm{pred}^{(i)} = [\bm{{x}}_\textrm{pred}^{(i)}, \bm{{y}}_\textrm{pred}^{(i)}]^\textrm{T}$}, where $\bm{{x}}_\textrm{pred}^{(i)}, \bm{{y}}_\textrm{pred}^{(i)} \in \mathbb{R}^{H_\textrm{pred}}$ and $H_\textrm{pred}=T_\textrm{pred}\cdot f$.
\subsection{Datasets}
For the experiments, the publicly available highway datasets highD~\cite{highD} and NGSIM~\cite{ngsim.2017} are used due to their extent of application-oriented scenarios. 

\textbf{highD:} The dataset provides traffic scenarios collected at six different highways in Germany, varying by the number of lanes and speed limits. In total, the dataset consists of $\SI{16.5}{\hour}$ of measurements captured with a sample rate of $f = \SI{25}{\hertz}$.

\textbf{NGSIM:} The dataset includes traffic recordings from the US highways US-101 and I-80, captured with a sample rate of $f = \SI{10}{\hertz}$. For each highway, a total recording time of $\SI{45}{\min}$ with different traffic conditions is provided.
 
\begin{table}[t!]
	\centering
	\vspace{7pt}
	\begin{tabular}{c|c|c|c|c|}
		\multicolumn{1}{c|}{Ablation}&
		\multicolumn{2}{c|}{highD} & 
		\multicolumn{2}{c|}{NGSIM} \\
		$q$ &ADE [\SI{}{\meter}] 
		& FDE [\SI{}{\meter}]  & ADE [\SI{}{\meter}]  & FDE [\SI{}{\meter}]  \\
		\hline
		\hline
		30 	    	& 0.80 & 2.09 & 10.58 & 23.32  \\	\hline 
		40 	   	    & {0.73} & \textbf{1.80} & 9.68 & 21.05 \\	\hline
		50  	    	& 0.89 & 1.82 & \textbf{8.91} & \textbf{18.86}  \\	\hline 
		80  	    	& \textbf{0.72 }&\textbf{ 1.80} & - & -  \\	\hline 
		125  			& 0.81 & 1.87 & - & -  \\	\hline 
	\end{tabular}%
	\caption{Prediction performance of GFTNNv2 with varying settings of the parameter $q$, which limits the number of predicted frequencies. Based on the traffic observations of \SI{3}{\second}, each model has to predict the trajectory of one selected agent for the next \SI{5}{\second}. For the NGSIM dataset, experiments for $q>50$ are not possible since $0 < q\leq H_\textrm{pred}=50$.}
	\label{tab:ablations}
	\vspace{-12pt}
\end{table}
\begin{table*}[t!]
	\centering
	\vspace{7pt}
	\begin{tabular}{c|c|c|c|c|}
		\multirow{2}{*}{Architecture}&
		\multicolumn{2}{c|}{highD} & 
		\multicolumn{2}{c|}{NGSIM} \\
		&ADE [\SI{}{\meter}]@\SI{5}{\second} 
		& FDE [\SI{}{\meter}]@\SI{5}{\second} & ADE [\SI{}{\meter}]@\SI{5}{\second} & FDE [\SI{}{\meter}]@\SI{5}{\second}  \\
		\hline
		\hline
		\textbf{GFTNNv2$-$50}	    	&  0.89 & 1.82 & {8.91} & {18.86} \\	\hline
		\textbf{GFTNNv2$-$80}	    	&  \textbf{0.72} & \textbf{1.80} & - & -  \\	\hline 
		{GFTNN \cite{Neumeier.2022}}	     	& {1.04} & {2.41} & $\mathbf{3.08}$ & \textbf{7.31}  \\	\hline 
		HSTA \cite{HSTA.2021}        	& 2.18 & 4.56 & 10.36 & 22.84	\\	\hline
		DVAE\cite{Neumeier.2021}			 	& 4.21 & 4.38 & 3.45 & 7.36	\\	\hline
		CS-LSTM	 \cite{Deo.2018}	 	& 2.88 & 5.71 & 8.65 & 16.93\\	\hline 
		MHA-LSTM(+f) \cite{Messaoud.2021} & 2.58 & 5.44 & 13.10 & 27.45	\\	\hline 
		Two-channel \cite{Mo.08.07.2021}  & 2.97 & 6.30 & 6.55 & 14.13	\\	\hline 
		RA-GAT \cite{RAGAT.2021} & 3.46 & 6.93 & 7.05 & 15.49 \\ \hline 
		\hline
	\end{tabular}%
	\caption{Performance of different state-of-the-art architectures for trajectory predictions on highway scenarios. Based on the traffic observations of \SI{3}{\second}, each model has to predict the trajectory of one selected agent for the next \SI{5}{\second}. The used metrics for evaluation are ADE and FDE.}
	\label{tab:comparison}
	\vspace{-8pt}
\end{table*}
As both datasets naturally have a huge imbalance of scenarios \cite{RAGAT.2021}, each dataset is pre-processed in such a fashion that they hold an equal distribution of scenario types. The resulting data format aligns with the problem definition explained beforehand. Training based on highly imbalanced dataset typically results in high prediction accuracy for the majority class (lane keeping maneuvers) but the model fails to capture minority classes (lane changes). 
From the highD/NGSIM dataset 9000/1100 highway scenarios are extracted, that represent an equal distribution of the maneuvers keep lane, lane change to the right and lane change to the left. Due to a limited number of lane change scenarios in the NGSIM dataset, the amount of total scenarios is reduced. 

\subsection{Implementation Details and Ablation Study}
The train-test split is set to $\SI{70}{\percent}-\SI{30}{\percent}$ for all runs. Optimization is done using Adam with a learning rate of $lr = 1e-4$ based on 60 (highD) or 120 (NGSIM) epochs, respectively. The loss function $\mathcal{L}$ used for training is the Mean Squared Error (MSE)
\begin{equation}
	\mathcal{L}= \text{MSE}(\bm{{x}}_\textrm{pred}^{(i)}, \bm{x}) + \text{MSE}(\bm{{y}}_\textrm{pred}^{(i)}, \bm{y}),
\end{equation}
where $\bm{x},\bm{y} \in \mathbb{R}^{H_\textrm{pred}}$ is the ground truth trajectory.
The benchmarked models are trained and evaluated on the single agent trajectory prediction task using the same dataset. The implementations of baseline models, for which the code was not made publicly available, are based on the information provided in the related publication. 

For GFTNNv2, the following ablations are evaluated. 

\textbf{GFTNNv2$-q$:} The prediction module of GFTNNv2 predicts information of the \mbox{$q \in \{30, 40, 50, 80, 125\}$} lowest temporal eigenvalues (frequencies) of the graph spectral scenario representation. 

\subsection{Evaluation and Results}
GFTNNv2 is compared with state-of-the-art trajectory prediction methods including conventional recurrent models (e.\,g., CS-LSTM \cite{Deo.2018}, MHA-LSTM(+f) \cite{Messaoud.2021}), novel graph-based approaches (e.\,g., HSTA \cite{HSTA.2021}, Two-Channel GNN \cite{Mo.08.07.2021}, RA-GAT \cite{RAGAT.2021}) and the previously introduced GFTNN\cite{Neumeier.2022}. 

The evaluation process employs two commonly used metrics in the field of ML-based trajectory prediction: average displacement error and final displacement error.
\subsubsection{Average Displacment Error (ADE)} The average of the Root Mean Squared Error (RMSE) between the trajectory prediction and the ground truth of every time step.
\subsubsection{Final Displacment Error (FDE)} The RMSE between the trajectory prediction and the ground truth at the last prediction step.

In Table~\ref{tab:ablations}, the prediction performance due to different GFTNNv2 settings is evaluated. This study enables exploring the impact of the selection of the hyper-parameter $q$. Parameter $q$ defines the number of predicted frequencies and the higher its value, the more periodic function terms (eigenvectors) are used to construct the trajectory prediction. Due to the definition of the GFT, $q$ is restricted to $q\leq  H_\textrm{pred}$. Thus, for the highD dataset experiments are limited to \mbox{$q\leq H_\textrm{pred}=125$} and for the NGSIM dataset to \mbox{$q\leq H_\textrm{pred}=50$}. As can be seen from the table, insufficiently accounting for frequencies ($q=30$) in the prediction model can lead to poor performance in making accurate predictions. On the contrary, considering too many frequencies (\mbox{$q=125$}) and period function terms respectively can also negatively impact the prediction performance. Using particularly high frequency terms may not have a meaningful impact on the accuracy of predictions, and may instead introduce unnecessary noise. Also, considering more frequency terms increases prediction complexity and more model parameters have to be learned.
Furthermore, Table~\ref{tab:ablations} indicates, that certain definitions of $q$ may have a negative impact on prediction performance. On the highD dataset, the parameter setting \mbox{$q=50$} leads to a noticeably less accurate trajectory prediction according to the ADE metric. 
The best performing model setting of GFTNNv2 is GFTNNv2-$80$ for the highD dataset and GFTNNv2-$50$ for the NGSIM. For benchmarking the proposed architecture with baseline models these GFTNNv2 settings are used.

Table~\ref{tab:comparison} assesses the prediction performance of the selected GFTNNv2 realizations and the considered state-of-the-art models. On the highD dataset, the proposed GFTNNv2-80 shows a superior performance w.\,r.\,t. the benchmarked models. GFTNNv2-80 demonstrates a significant improvement in both evaluation metrics, achieving a \SI{29}{\percent} (ADE) and \SI{25}{\percent} (FDE) increase compared to the best-performing benchmark model. Although GFTNNv2-50 does not perform as competitive as GFTNNv2-80 on highD, it also outperforms the baseline models in both of the used evaluation metrics. On the NGSIM dataset, the GFTNNv2 realizations do not achieve competitive performance w.\,r.\,t. GFTNN\cite{Neumeier.2022}, which provides most accurate results in the prediction task. When analyzing Table~\ref{tab:comparison}, however, it is conspicuous that all benchmarked models except DVAE\cite{Neumeier.2021} show a decline in prediction performance on NGSIM dataset in comparison to highD dataset. 
One possible reason for this could be that the trajectories in NGSIM are generally more complex and challenging to predict. Another possible reason could be that there is too little data available in the pre-processed NGSIM dataset. This makes it difficult for the models to learn to generalize, and as a result, models tend to overfit to the training data. As mentioned initially, the pre-processed NGSIM dataset consists of only 1100 scenarios in total compared to 9000 available scenarios in the highD dataset. Due to the significant imbalance of scenarios in the original dataset\cite{RAGAT.2021}, it is not reasonable to train on the entire dataset. 
In fact, it is noteworthy that the two models that perform significantly better than all other prediction models on the NGSIM dataset (GFTNN and DVAE), use a descriptive decoder \cite{Neumeier.2021}. The descriptive decoder is a physics-informed network. Such architectures are known to generalize well even in cases of limited data availability. The descriptive decoder introduces prior knowledge of vehicle kinematic and thereby limits the space of admissible solutions. This supports the assumption that the used NGSIM dataset might be too complex and small to allow generalization for common statistical learning models. If sufficient data is available, however, the proposed GFTNNv2 shows superior performance and is the preferred choice. 
\section{Conclusion}
This work introduces GFTNNv2, a statistical learning network for vehicle trajectory predictions, leveraging the power of deep learning and graph theory. Based on the concept of the multidimensional GFT, GFTNNv2 learns to predict future trajectories in the graph spectral domain. By using a spatio-temporal graph to represent a traffic scenario and applying multidimensional GFT, an effective representation of the scenario in terms of time, space, and relations is obtained. The resulting graph spectral representation of the traffic scenario is the basis for the prediction and allows the introduction of filtering operations to reduce the learning task complexity. As shown in this work, however, filtering in the spatial axis of the graph spectral domain is not reasonable as it undermines the traffic context. Despite its simple architecture, GFTNNv2 performs $\sim \SI{25}{\percent}$ better than the top-performing state-of-the-art approach when predicting vehicle trajectories on highways based on the highD dataset.
In future studies, the research focus will be on thoroughly analyzing why there is a significant performance variability among datasets and devising strategies to improve the overall prediction performance.
{
	\bibliographystyle{IEEEtran}
	\bibliography{ref/ref_gilbertSPEC.bib}
}
\end{document}

%% file: fig/GraphBuildup_8.pdf_tex
\begingroup%
  \makeatletter%
  \providecommand\color[2][]{%
    \errmessage{(Inkscape) Color is used for the text in Inkscape, but the package 'color.sty' is not loaded}%
    \renewcommand\color[2][]{}%
  }%
  \providecommand\transparent[1]{%
    \errmessage{(Inkscape) Transparency is used (non-zero) for the text in Inkscape, but the package 'transparent.sty' is not loaded}%
    \renewcommand\transparent[1]{}%
  }%
  \providecommand\rotatebox[2]{#2}%
  \newcommand*\fsize{\dimexpr\f@size pt\relax}%
  \newcommand*\lineheight[1]{\fontsize{\fsize}{#1\fsize}\selectfont}%
  \ifx\svgwidth\undefined%
    \setlength{\unitlength}{358.59643601bp}%
    \ifx\svgscale\undefined%
      \relax%
    \else%
      \setlength{\unitlength}{\unitlength * \real{\svgscale}}%
    \fi%
  \else%
    \setlength{\unitlength}{\svgwidth}%
  \fi%
  \global\let\svgwidth\undefined%
  \global\let\svgscale\undefined%
  \makeatother%
  \begin{picture}(1,0.56549424)%
    \lineheight{1}%
    \setlength\tabcolsep{0pt}%
    \put(0,0){\includegraphics[width=\unitlength,page=1]{fig/GraphBuildup_8.pdf}}%
    \put(0.1488322,0.02496628){\color[rgb]{0,0,0}\makebox(0,0)[lt]{\lineheight{1.25}\smash{\begin{tabular}[t]{l}0\end{tabular}}}}%
    \put(0.1488322,0.06727372){\color[rgb]{0,0,0}\makebox(0,0)[lt]{\lineheight{1.25}\smash{\begin{tabular}[t]{l}1\end{tabular}}}}%
    \put(0.1488322,0.32111773){\color[rgb]{0,0,0}\makebox(0,0)[lt]{\lineheight{1.25}\smash{\begin{tabular}[t]{l}7\end{tabular}}}}%
    \put(0.1488322,0.36342517){\color[rgb]{0,0,0}\makebox(0,0)[lt]{\lineheight{1.25}\smash{\begin{tabular}[t]{l}8\end{tabular}}}}%
    \put(0.1488322,0.10958116){\color[rgb]{0,0,0}\makebox(0,0)[lt]{\lineheight{1.25}\smash{\begin{tabular}[t]{l}2\end{tabular}}}}%
    \put(0.1488322,0.15188833){\color[rgb]{0,0,0}\makebox(0,0)[lt]{\lineheight{1.25}\smash{\begin{tabular}[t]{l}3\end{tabular}}}}%
    \put(0.1488322,0.19419577){\color[rgb]{0,0,0}\makebox(0,0)[lt]{\lineheight{1.25}\smash{\begin{tabular}[t]{l}4\end{tabular}}}}%
    \put(0.1488322,0.23650309){\color[rgb]{0,0,0}\makebox(0,0)[lt]{\lineheight{1.25}\smash{\begin{tabular}[t]{l}5\end{tabular}}}}%
    \put(0.1488322,0.27881041){\color[rgb]{0,0,0}\makebox(0,0)[lt]{\lineheight{1.25}\smash{\begin{tabular}[t]{l}6\end{tabular}}}}%
    \put(0.05111396,0.14677086){\color[rgb]{0,0,0}\makebox(0,0)[lt]{\lineheight{1.25}\smash{\begin{tabular}[t]{l}t\end{tabular}}}}%
    \put(0,0){\includegraphics[width=\unitlength,page=2]{fig/GraphBuildup_8.pdf}}%
    \put(0.2588067,0.14790166){\color[rgb]{0,0,0}\makebox(0,0)[lt]{\lineheight{1.25}\smash{\begin{tabular}[t]{l}y\end{tabular}}}}%
    \put(0.41641032,0){\color[rgb]{0,0,0}\makebox(0,0)[lt]{\lineheight{1.25}\smash{\begin{tabular}[t]{l}x\end{tabular}}}}%
    \put(0,0){\includegraphics[width=\unitlength,page=3]{fig/GraphBuildup_8.pdf}}%
    \put(0.81333684,0.00199973){\color[rgb]{0,0,0}\makebox(0,0)[lt]{\lineheight{1.25}\smash{\begin{tabular}[t]{l}x\end{tabular}}}}%
    \put(0.65851963,0.14132317){\color[rgb]{0,0,0}\makebox(0,0)[lt]{\lineheight{1.25}\smash{\begin{tabular}[t]{l}t\end{tabular}}}}%
    \put(0.71496075,0.10896685){\color[rgb]{0,0,0}\makebox(0,0)[lt]{\lineheight{1.25}\smash{\begin{tabular}[t]{l}y\end{tabular}}}}%
    \put(0,0){\includegraphics[width=\unitlength,page=4]{fig/GraphBuildup_8.pdf}}%
    \put(0.30929605,0.11319807){\color[rgb]{0,0,0}\makebox(0,0)[lt]{\lineheight{1.25}\smash{\begin{tabular}[t]{l}8\end{tabular}}}}%
    \put(0.37381271,0.05695283){\color[rgb]{0,0,0}\makebox(0,0)[lt]{\lineheight{1.25}\smash{\begin{tabular}[t]{l}7\end{tabular}}}}%
    \put(0.57470831,0.11610738){\color[rgb]{0,0,0}\makebox(0,0)[lt]{\lineheight{1.25}\smash{\begin{tabular}[t]{l}5\end{tabular}}}}%
    \put(0.50766304,0.05687672){\color[rgb]{0,0,0}\makebox(0,0)[lt]{\lineheight{1.25}\smash{\begin{tabular}[t]{l}6\end{tabular}}}}%
    \put(0.57744749,0.21388641){\color[rgb]{0,0,0}\makebox(0,0)[lt]{\lineheight{1.25}\smash{\begin{tabular}[t]{l}4\end{tabular}}}}%
    \put(0.31282796,0.21408688){\color[rgb]{0,0,0}\makebox(0,0)[lt]{\lineheight{1.25}\smash{\begin{tabular}[t]{l}1\end{tabular}}}}%
    \put(0.4393934,0.20711647){\color[rgb]{0,0,0}\makebox(0,0)[lt]{\lineheight{1.25}\smash{\begin{tabular}[t]{l}0\end{tabular}}}}%
    \put(0.50799836,0.29576181){\color[rgb]{0,0,0}\makebox(0,0)[lt]{\lineheight{1.25}\smash{\begin{tabular}[t]{l}3\end{tabular}}}}%
    \put(0.37821102,0.29562817){\color[rgb]{0,0,0}\makebox(0,0)[lt]{\lineheight{1.25}\smash{\begin{tabular}[t]{l}2\end{tabular}}}}%
    \put(0,0){\includegraphics[width=\unitlength,page=5]{fig/GraphBuildup_8.pdf}}%
    \put(0.13643844,0.4214765){\color[rgb]{0,0,0}\makebox(0,0)[t]{\lineheight{1.25}\smash{\begin{tabular}[t]{c}$\mathcal{G}_1$\end{tabular}}}}%
    \put(0.45146801,0.35113697){\color[rgb]{0,0,0}\makebox(0,0)[t]{\lineheight{1.25}\smash{\begin{tabular}[t]{c}$\mathcal{G}_2$\end{tabular}}}}%
    \put(0.87100425,0.54188872){\color[rgb]{0,0,0}\makebox(0,0)[t]{\lineheight{1.25}\smash{\begin{tabular}[t]{c}$\mathcal{G}_1 \square \mathcal{G}_2$\end{tabular}}}}%
  \end{picture}%
\endgroup%

%% file: tikz/pred_model.tex
\pgfplotsset{
	matrix plot/.style={
		axis on top,
		clip marker paths=true,
		scale only axis,
		height=\matrixrows/\matrixcols*\pgfkeysvalueof{/pgfplots/width},
		enlarge x limits={rel=0.5/\matrixcols},
		enlarge y limits={rel=0.5/\matrixrows},
		scatter/use mapped color={draw=mapped color, fill=mapped color},
		scatter,
		point meta=explicit,
		mark=square*,
		cycle list={
			mark size=0.5*\pgfkeysvalueof{/pgfplots/width}/\matrixcols
		}
	},
	matrix rows/.store in=\matrixrows,
	matrix rows=8,
	matrix cols/.store in=\matrixcols,
	matrix cols=10
}
\begin{tikzpicture}
\def \radS {2.5mm}
\coordinate (xn1) at(-2.9,0);
\coordinate (x0) at (0,0);
\coordinate (g0) at (-0.7, +0.1);
\coordinate (x1) at (1, 0.8);
\coordinate (x2) at (-1, -0.5);
\coordinate (x3) at (1.3, 0.05);

\coordinate (c0) at ($(x0) + (3, 0)$);
\def \rectSize {2.6cm}
\def \descriptiveSize {3cm}
\coordinate (c0) at ($(x0) + (3, 0)$);
\coordinate (c1) at ($(c0) + (0.5*\rectSize, 0)$);

\coordinate (nn0) at ($(c0)+ (0.95*\rectSize,0)$);
\coordinate (nnOUT) at ($(nn0) + (3.5,0)$);

\node[rectangle,
very thick,
draw= gray, 
minimum width = 0.97\columnwidth, 
minimum height = 4cm,
draw opacity = 0.4,
anchor=center,
label={[anchor=south]below:\textsc{\small Observation Module}},
align = right ] (obs) at ($(x0)+ (1.55cm,0)$) {};

\node[rectangle,
very thick,
draw= gray, 
minimum width = 0.85\columnwidth, 
minimum height = 4cm,
draw opacity = 0.4,
label={[anchor=south]below:\textsc{\small Prediction Module}},
anchor=west] (pred) at ($(obs) + (\columnwidth, 0) -(4.4cm,0)$) {};

\draw[-, white] (xn1) -- ++ (-0.0001,0);
\draw[ -, thick, fill = white] (x0) -- (x1);
\draw[ -, thick, fill = white] (x0) -- (x2);
\draw[ -, thick, fill = white] (x0) -- (x3);

\draw[ -, thick, fill = white] (x0) -- (x1);
\draw[ -, thick, fill = white] (x0) -- (x2);
\draw[ -, thick, fill = white] (x0) -- (x3);

\draw [ ->, very thick , fill = white] ($(x0) + (1.6,0)$) -- (c0) node [pos=.5, above] (TextNode1) {\textit{GFT}};
\node [anchor=center, minimum width=0.5cm, scale=0.6] at (g0) {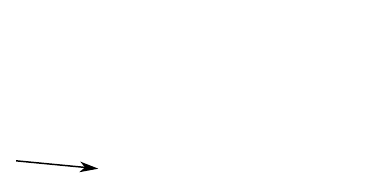};
\node[anchor=center,inner sep=0] at (x0) {\includegraphics[width=.5cm]{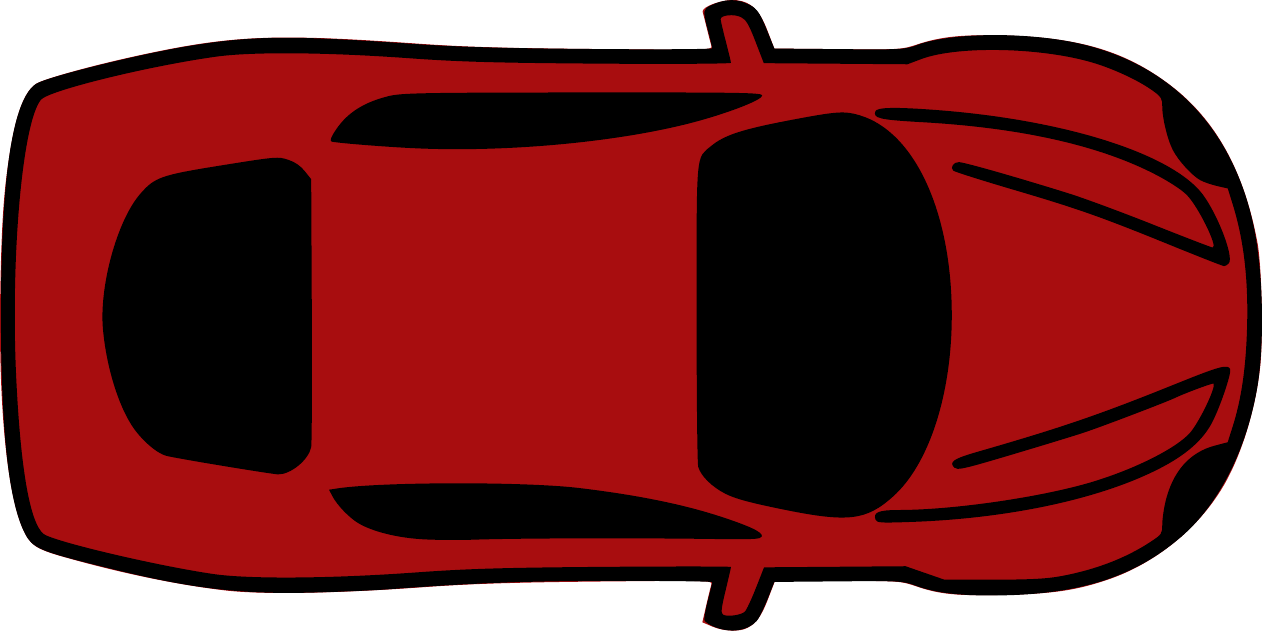}};
\node[anchor=center,inner sep=0] at (x1) {\includegraphics[width=.5cm]{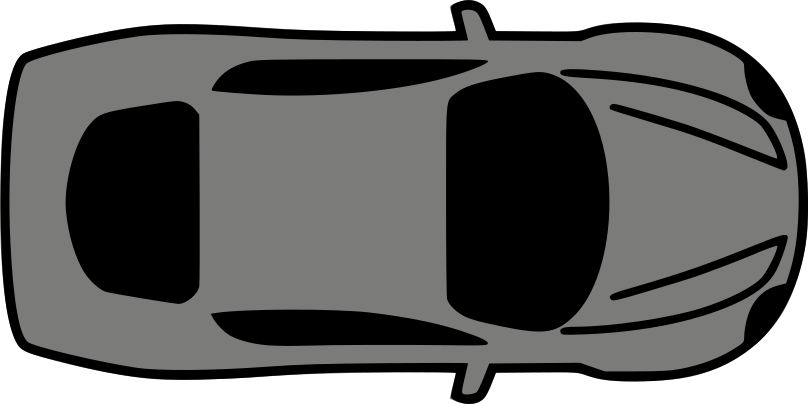}};
\node[anchor=center,inner sep=0] at (x2) {\includegraphics[width=.5cm]{tikz/fig/vehicle_gray}};
\node[anchor=center,inner sep=0] at (x3) {\includegraphics[width=.5cm]{tikz/fig/vehicle_gray}};

\node [anchor= west, draw,  dotted, thick, shape=rectangle, minimum width=\rectSize, minimum height=\rectSize] at (c0) {};
\begin{axis}[
at={($(c1) + (3pt, 5pt)$)},
anchor=center,
width=1.8cm, 
xticklabels={,,},
yticklabels={,,},
height = 2cm,
matrix plot,
colormap/viridis,
]
\addplot table [meta=funceval] {
ppm popsize timessolved funceval
0.005   100 30  5.2312e+06
0.105   100 30  1.6282e+06
0.205   100 29  2.0174e+06
0.305   100 30  1.66e+06
0.405   100 28  2.639e+06
0.505   100 24  4.404e+06
0.605   100 16  7.9352e+06
0.705   100 8   1.1446e+07
0.805   100 6   1.2217e+07
0.905   100 9   1.0817e+07

0.005   150 30  8.688e+06
0.105   150 30  2.1773e+06
0.205   150 30  2.2315e+06
0.305   150 30  2.401e+06
0.405   150 30  2.6153e+06
0.505   150 26  5.426e+06
0.605   150 21  8.7765e+06
0.705   150 23  7.0593e+06
0.805   150 16  1.1344e+07
0.905   150 14  1.2771e+07

0.005   200 30  6.6252e+06
0.105   200 30  3.0197e+06
0.205   200 30  2.9699e+06
0.305   200 30  3.1497e+06
0.405   200 30  3.5268e+06
0.505   200 30  3.7531e+06
0.605   200 27  6.1353e+06
0.705   200 24  8.3013e+06
0.805   200 21  1.0925e+07
0.905   200 17  1.4032e+07

0.005   250 30  8.1181e+06
0.105   250 30  3.5341e+06
0.205   250 30  3.8653e+06
0.305   250 30  3.7785e+06
0.405   250 30  4.3767e+06
0.505   250 30  4.3917e+06
0.605   250 29  5.3275e+06
0.705   250 29  4.6693e+06
0.805   250 23  1.0961e+07
0.905   250 25  8.02e+06

0.005   300 30  8.5783e+06
0.105   300 30  4.0582e+06
0.205   300 30  4.263e+06
0.305   300 30  4.692e+06
0.405   300 30  5.1478e+06
0.505   300 30  5.4769e+06
0.605   300 29  6.3917e+06
0.705   300 26  9.4055e+06
0.805   300 26  8.7111e+06
0.905   300 18  1.9458e+07

0.005   350 30  1.0232e+07
0.105   350 30  4.5218e+06
0.205   350 30  4.9123e+06
0.305   350 30  5.3983e+06
0.405   350 30  5.9751e+06
0.505   350 30  6.3135e+06
0.605   350 30  5.7111e+06
0.705   350 30  4.6045e+06
0.805   350 27  8.4856e+06
0.905   350 24  1.2849e+07

0.005   400 30  8.3002e+06
0.105   400 30  5.2684e+06
0.205   400 30  5.5534e+06
0.305   400 30  6.1292e+06
0.405   400 30  6.6702e+06
0.505   400 30  7.2192e+06
0.605   400 30  6.9306e+06
0.705   400 30  5.5206e+06
0.805   400 30  4.0857e+06
0.905   400 29  5.1603e+06

0.005   450 30  5.0286e+06
0.105   450 30  5.8948e+06
0.205   450 30  5.9761e+06
0.305   450 30  5.5749e+06
0.405   450 30  7.6008e+06
0.505   450 30  5.2071e+06
0.605   450 30  7.7313e+06
0.705   450 30  6.0063e+06
0.805   450 29  6.7191e+06
0.905   450 28  7.9783e+06
};
\end{axis}    
\begin{axis}[
at={($(c1) + (1pt, 1.5pt)$)},
anchor=center,
width=1.8cm, 
xticklabels={,,},
yticklabels={,,},
height = 2cm,
matrix plot,
colormap/viridis,
]
\addplot table [meta=funceval] {
	ppm popsize timessolved funceval
	0.005   100 30  5.2312e+06
	0.105   100 30  1.6282e+06
	0.205   100 29  2.0174e+06
	0.305   100 30  1.66e+06
	0.405   100 28  2.639e+06
	0.505   100 24  4.404e+06
	0.605   100 16  7.9352e+06
	0.705   100 8   1.1446e+07
	0.805   100 6   1.2217e+06
	0.905   100 9   1.0817e+06
	
	0.005   150 30  8.688e+06
	0.105   150 30  2.1773e+06
	0.205   150 30  2.2315e+06
	0.305   150 30  2.401e+06
	0.405   150 30  2.6153e+06
	0.505   150 26  5.426e+06
	0.605   150 21  8.7765e+06
	0.705   150 23  7.0593e+06
	0.805   150 16  1.1344e+07
	0.905   150 14  1.2771e+06
	
	0.005   200 30  6.6252e+06
	0.105   200 30  3.0197e+06
	0.205   200 30  2.9699e+06
	0.305   200 30  3.1497e+06
	0.405   200 30  3.5268e+06
	0.505   200 30  3.7531e+06
	0.605   200 27  6.1353e+06
	0.705   200 24  8.3013e+06
	0.805   200 21  1.0925e+06
	0.905   200 17  1.4032e+06
	
	0.005   250 30  8.1181e+06
	0.105   250 30  3.5341e+06
	0.205   250 30  3.8653e+06
	0.305   250 30  3.7785e+06
	0.405   250 30  4.3767e+06
	0.505   250 30  4.3917e+06
	0.605   250 29  5.3275e+06
	0.705   250 29  4.6693e+06
	0.805   250 23  1.0961e+07
	0.905   250 25  8.02e+06
	
	0.005   300 30  8.5783e+06
	0.105   300 30  4.0582e+06
	0.205   300 30  4.263e+06
	0.305   300 30  4.692e+06
	0.405   300 30  5.1478e+06
	0.505   300 30  5.4769e+06
	0.605   300 29  6.3917e+06
	0.705   300 26  9.4055e+06
	0.805   300 26  8.7111e+06
	0.905   300 18  8.9458e+06
	
	0.005   350 30  1.0232e+07
	0.105   350 30  4.5218e+06
	0.205   350 30  4.9123e+06
	0.305   350 30  5.3983e+06
	0.405   350 30  5.9751e+06
	0.505   350 30  6.3135e+06
	0.605   350 30  5.7111e+06
	0.705   350 30  4.6045e+06
	0.805   350 27  8.4856e+06
	0.905   350 24  4.2849e+06
	
	0.005   400 30  8.3002e+06
	0.105   400 30  5.2684e+06
	0.205   400 30  5.5534e+06
	0.305   400 30  6.1292e+06
	0.405   400 30  6.6702e+06
	0.505   400 30  7.2192e+06
	0.605   400 30  6.9306e+06
	0.705   400 30  5.5206e+06
	0.805   400 30  4.0857e+06
	0.905   400 29  5.1603e+06
	
	0.005   450 30  1.0286e+07
	0.105   450 30  5.8948e+06
	0.205   450 30  5.9761e+06
	0.305   450 30  6.5749e+06
	0.405   450 30  7.6008e+06
	0.505   450 30  8.2071e+06
	0.605   450 30  7.7313e+06
	0.705   450 30  6.0063e+06
	0.805   450 29  6.7191e+06
	0.905   450 28  7.9783e+06
};
\end{axis}    
\begin{axis}[
at={($(c1) + (-1pt, -1.5pt)$)},
anchor=center,
width=1.8cm, 
xticklabels={,,},
yticklabels={,,},
height = 2cm,
matrix plot,
colormap/viridis,
]
\addplot table [meta=funceval] {
	ppm popsize timessolved funceval
	0.005   100 30  2.2312e+06
	0.105   100 30  1.6282e+06
	0.205   100 29  2.0174e+06
	0.305   100 30  1.66e+06
	0.405   100 28  2.639e+06
	0.505   100 24  4.404e+06
	0.605   100 16  7.9352e+06
	0.705   100 8   1.1446e+06
	0.805   100 6   1.2217e+06
	0.905   100 9   1.817e+06
	
	0.005   150 30  8.688e+06
	0.105   150 30  2.1773e+06
	0.205   150 30  2.2315e+06
	0.305   150 30  2.401e+06
	0.405   150 30  2.6153e+06
	0.505   150 26  5.426e+06
	0.605   150 21  8.7765e+06
	0.705   150 23  7.0593e+06
	0.805   150 16  1.1344e+07
	0.905   150 14  1.0771e+07
	
	0.005   200 30  6.6252e+06
	0.105   200 30  3.0197e+06
	0.205   200 30  2.9699e+06
	0.305   200 30  3.1497e+06
	0.405   200 30  3.5268e+06
	0.505   200 30  3.7531e+06
	0.605   200 27  6.1353e+06
	0.705   200 24  8.3013e+06
	0.805   200 21  8.0925e+06
	0.905   200 17  8.4032e+06
	
	0.005   250 30  8.1181e+06
	0.105   250 30  3.5341e+06
	0.205   250 30  3.8653e+06
	0.305   250 30  3.7785e+06
	0.405   250 30  4.3767e+06
	0.505   250 30  4.3917e+06
	0.605   250 29  5.3275e+06
	0.705   250 29  4.6693e+06
	0.805   250 23  7.0961e+06
	0.905   250 25  7.02e+06
	
	0.005   300 30  5.5783e+06
	0.105   300 30  4.0582e+06
	0.205   300 30  4.263e+06
	0.305   300 30  4.692e+06
	0.405   300 30  3.1478e+06
	0.505   300 30  3.4769e+06
	0.605   300 29  1.3917e+06
	0.705   300 26  2.4055e+06
	0.805   300 26  4.7111e+06
	0.905   300 18  1.0458e+07
	
	0.005   350 30  1.0232e+07
	0.105   350 30  4.5218e+06
	0.205   350 30  4.9123e+06
	0.305   350 30  5.3983e+06
	0.405   350 30  5.9751e+06
	0.505   350 30  6.3135e+06
	0.605   350 30  5.7111e+06
	0.705   350 30  4.6045e+06
	0.805   350 27  8.4856e+06
	0.905   350 24  9.2849e+06
	
	0.005   400 30  8.3002e+06
	0.105   400 30  5.2684e+06
	0.205   400 30  5.5534e+06
	0.305   400 30  6.1292e+06
	0.405   400 30  6.6702e+06
	0.505   400 30  7.2192e+06
	0.605   400 30  6.9306e+06
	0.705   400 30  5.5206e+06
	0.805   400 30  4.0857e+06
	0.905   400 29  5.1603e+06
	
	0.005   450 30  8.0286e+06
	0.105   450 30  5.8948e+06
	0.205   450 30  5.9761e+06
	0.305   450 30  6.5749e+06
	0.405   450 30  7.6008e+06
	0.505   450 30  8.2071e+06
	0.605   450 30  7.7313e+06
	0.705   450 30  8.0063e+06
	0.805   450 29  8.7191e+06
	0.905   450 28  8.9783e+06
};
\end{axis}    

\begin{axis}[
at={($(c1) + (-3pt, -5pt)$)},
anchor=center,
width=1.8cm, 
xticklabels={,,},
yticklabels={,,},
height = 2cm,
matrix plot,
colormap/viridis,
]
\addplot table [meta=funceval] {
	ppm popsize timessolved funceval
	0.005   100 30  1.2312e+06
	0.105   100 30  1.6282e+06
	0.205   100 29  2.0174e+06
	0.305   100 30  2.66e+06
	0.405   100 28  2.639e+06
	0.505   100 24  3.404e+06
	0.605   100 16  4.9352e+06
	0.705   100 8   4.1446e+06
	0.805   100 6   8.2217e+06
	0.905   100 9   9.0817e+06
	
	0.005   150 30  1.688e+06
	0.105   150 30  2.1773e+06
	0.205   150 30  2.2315e+06
	0.305   150 30  2.401e+06
	0.405   150 30  2.6153e+06
	0.505   150 26  3.426e+06
	0.605   150 21  4.7765e+06
	0.705   150 23  6.0593e+06
	0.805   150 16  1.1344e+07
	0.905   150 14  1.2771e+07
	
	0.005   200 30  2.8252e+06
	0.105   200 30  3.0197e+06
	0.205   200 30  2.9699e+06
	0.305   200 30  3.1497e+06
	0.405   200 30  3.5268e+06
	0.505   200 30  3.7531e+06
	0.605   200 27  6.1353e+06
	0.705   200 24  8.3013e+06
	0.805   200 21  1.3925e+07
	0.905   200 17  1.4032e+07
	
	0.005   250 30  6.1181e+06
	0.105   250 30  5.5341e+06
	0.205   250 30  5.8653e+06
	0.305   250 30  4.7785e+06
	0.405   250 30  4.3767e+06
	0.505   250 30  4.3917e+06
	0.605   250 29  5.3275e+06
	0.705   250 29  4.6693e+06
	0.805   250 23  1.1961e+07
	0.905   250 25  1.22e+07
	
	0.005   300 30  8.5783e+06
	0.105   300 30  9.0582e+06
	0.205   300 30  9.263e+06
	0.305   300 30  7.692e+06
	0.405   300 30  5.1478e+06
	0.505   300 30  5.4769e+06
	0.605   300 29  6.3917e+06
	0.705   300 26  9.4055e+06
	0.805   300 26  9.9111e+06
	0.905   300 18  1.9458e+07
	
	0.005   350 30  1.0232e+07
	0.105   350 30  1.3218e+07
	0.205   350 30  9.9123e+06
	0.305   350 30  5.3983e+06
	0.405   350 30  5.9751e+06
	0.505   350 30  6.3135e+06
	0.605   350 30  5.7111e+06
	0.705   350 30  4.6045e+06
	0.805   350 27  8.4856e+06
	0.905   350 24  1.2849e+07
	
	0.005   400 30  8.3002e+06
	0.105   400 30  5.2684e+06
	0.205   400 30  5.5534e+06
	0.305   400 30  6.1292e+06
	0.405   400 30  6.6702e+06
	0.505   400 30  7.2192e+06
	0.605   400 30  6.9306e+06
	0.705   400 30  5.5206e+06
	0.805   400 30  4.0857e+06
	0.905   400 29  5.1603e+06
	
	0.005   450 30  1.0286e+07
	0.105   450 30  5.8948e+06
	0.205   450 30  5.9761e+06
	0.305   450 30  6.5749e+06
	0.405   450 30  7.6008e+06
	0.505   450 30  8.2071e+06
	0.605   450 30  7.7313e+06
	0.705   450 30  6.0063e+06
	0.805   450 29  6.7191e+06
	0.905   450 28  4.9783e+06
};
\end{axis}    

\node at ($(c1) + (0, -1.1)$) {\small $\mathbf{\hat{F}}_{\text{obs}}$};
\node at ($(nnOUT) + (0.5*\rectSize, -1.1)$) {\small $\mathbf{\hat{F}}_{\text{pred}}$};


\foreach \m/\l [count=\y] in {1, 2, 3,missing,4}
\node [anchor= west, every neuron/.try, neuron \m/.try] (input-\m) at ($(nn0)+(1,1.5-\y*0.5)$) {};

\foreach \m [count=\y] in {1,2, 3, missing,4}
\node [every neuron/.try, neuron \m/.try ] (hidden-\m) at ($(nn0)+(2, 1.5-\y*0.5)$) {};

\foreach \m [count=\y] in {1,2, missing,3}
\node [every neuron/.try, neuron \m/.try ] (output-\m) at ($(nn0)+(2.8, 1.2-\y*0.5)$) {};

\foreach \l [count=\i] in {1, 2, 3, Z}
\draw [<-] (input-\i) -- ++(-1,0)
node [above, midway] {$s_\l$};
%

\foreach \l [count=\i] in {1, 2, n}
\draw [->] (output-\i) -- ++ (0.65,0) node[] {};
\node [anchor= west, align=center, draw,  dotted, thick, shape=rectangle, minimum width=\rectSize, minimum height=\rectSize] at (nnOUT) {};

\begin{axis}[
at={($(nnOUT) +(0.5*\rectSize,0) + (1pt, 1.5pt)$)},
anchor=center,
width=1.8cm, 
xticklabels={,,},
yticklabels={,,},
height = 2cm,
matrix plot,
colormap/viridis,
]
\addplot table [meta=funceval] {
	ppm popsize timessolved funceval
	0.005   100 30  5.2312e+06
	0.105   100 30  1.6282e+06
	0.205   100 29  2.0174e+06
	0.305   100 30  1.66e+06
	0.405   100 28  2.639e+06
	0.505   100 24  4.404e+06
	0.605   100 16  7.9352e+06
	0.705   100 8   1.1446e+07
	0.805   100 6   1.2217e+06
	0.905   100 9   1.0817e+06
	
	0.005   150 30  8.688e+06
	0.105   150 30  2.1773e+06
	0.205   150 30  2.2315e+06
	0.305   150 30  2.401e+06
	0.405   150 30  2.6153e+06
	0.505   150 26  5.426e+06
	0.605   150 21  8.7765e+06
	0.705   150 23  7.0593e+06
	0.805   150 16  1.1344e+07
	0.905   150 14  1.2771e+06
	
	0.005   200 30  6.6252e+06
	0.105   200 30  3.0197e+06
	0.205   200 30  2.9699e+06
	0.305   200 30  3.1497e+06
	0.405   200 30  3.5268e+06
	0.505   200 30  3.7531e+06
	0.605   200 27  6.1353e+06
	0.705   200 24  8.3013e+06
	0.805   200 21  1.0925e+06
	0.905   200 17  1.4032e+06
	
	0.005   250 30  8.1181e+06
	0.105   250 30  3.5341e+06
	0.205   250 30  3.8653e+06
	0.305   250 30  3.7785e+06
	0.405   250 30  4.3767e+06
	0.505   250 30  4.3917e+06
	0.605   250 29  5.3275e+06
	0.705   250 29  4.6693e+06
	0.805   250 23  1.0961e+07
	0.905   250 25  8.02e+06
	
	0.005   300 30  8.5783e+06
	0.105   300 30  4.0582e+06
	0.205   300 30  4.263e+06
	0.305   300 30  4.692e+06
	0.405   300 30  5.1478e+06
	0.505   300 30  5.4769e+06
	0.605   300 29  6.3917e+06
	0.705   300 26  9.4055e+06
	0.805   300 26  8.7111e+06
	0.905   300 18  8.9458e+06
	
	0.005   350 30  1.0232e+07
	0.105   350 30  4.5218e+06
	0.205   350 30  4.9123e+06
	0.305   350 30  5.3983e+06
	0.405   350 30  5.9751e+06
	0.505   350 30  6.3135e+06
	0.605   350 30  5.7111e+06
	0.705   350 30  4.6045e+06
	0.805   350 27  8.4856e+06
	0.905   350 24  4.2849e+06
	
	0.005   400 30  8.3002e+06
	0.105   400 30  5.2684e+06
	0.205   400 30  5.5534e+06
	0.305   400 30  6.1292e+06
	0.405   400 30  6.6702e+06
	0.505   400 30  7.2192e+06
	0.605   400 30  6.9306e+06
	0.705   400 30  5.5206e+06
	0.805   400 30  4.0857e+06
	0.905   400 29  5.1603e+06
	
	0.005   450 30  1.0286e+07
	0.105   450 30  5.8948e+06
	0.205   450 30  5.9761e+06
	0.305   450 30  6.5749e+06
	0.405   450 30  7.6008e+06
	0.505   450 30  8.2071e+06
	0.605   450 30  7.7313e+06
	0.705   450 30  6.0063e+06
	0.805   450 29  6.7191e+06
	0.905   450 28  7.9783e+06
};
\end{axis}    
\begin{axis}[
at={($(nnOUT) +(0.5*\rectSize,0) +  (-1pt, -1.5pt)$)},
anchor=center,
width=1.8cm, 
xticklabels={,,},
yticklabels={,,},
height = 2cm,
matrix plot,
colormap/viridis,
]
\addplot table [meta=funceval] {
	ppm popsize timessolved funceval
	0.005   100 30  2.2312e+06
	0.105   100 30  1.6282e+06
	0.205   100 29  2.0174e+06
	0.305   100 30  1.66e+06
	0.405   100 28  2.639e+06
	0.505   100 24  4.404e+06
	0.605   100 16  7.9352e+06
	0.705   100 8   1.1446e+06
	0.805   100 6   1.2217e+06
	0.905   100 9   1.817e+06
	
	0.005   150 30  8.688e+06
	0.105   150 30  2.1773e+06
	0.205   150 30  2.2315e+06
	0.305   150 30  2.401e+06
	0.405   150 30  2.6153e+06
	0.505   150 26  5.426e+06
	0.605   150 21  8.7765e+06
	0.705   150 23  7.0593e+06
	0.805   150 16  1.1344e+07
	0.905   150 14  1.0771e+07
	
	0.005   200 30  6.6252e+06
	0.105   200 30  3.0197e+06
	0.205   200 30  2.9699e+06
	0.305   200 30  3.1497e+06
	0.405   200 30  3.5268e+06
	0.505   200 30  3.7531e+06
	0.605   200 27  6.1353e+06
	0.705   200 24  8.3013e+06
	0.805   200 21  8.0925e+06
	0.905   200 17  8.4032e+06
	
	0.005   250 30  8.1181e+06
	0.105   250 30  3.5341e+06
	0.205   250 30  3.8653e+06
	0.305   250 30  3.7785e+06
	0.405   250 30  4.3767e+06
	0.505   250 30  4.3917e+06
	0.605   250 29  5.3275e+06
	0.705   250 29  4.6693e+06
	0.805   250 23  7.0961e+06
	0.905   250 25  7.02e+06
	
	0.005   300 30  5.5783e+06
	0.105   300 30  4.0582e+06
	0.205   300 30  4.263e+06
	0.305   300 30  4.692e+06
	0.405   300 30  3.1478e+06
	0.505   300 30  3.4769e+06
	0.605   300 29  1.3917e+06
	0.705   300 26  2.4055e+06
	0.805   300 26  4.7111e+06
	0.905   300 18  1.0458e+07
	
	0.005   350 30  1.0232e+07
	0.105   350 30  4.5218e+06
	0.205   350 30  4.9123e+06
	0.305   350 30  5.3983e+06
	0.405   350 30  5.9751e+06
	0.505   350 30  6.3135e+06
	0.605   350 30  5.7111e+06
	0.705   350 30  4.6045e+06
	0.805   350 27  8.4856e+06
	0.905   350 24  9.2849e+06
	
	0.005   400 30  8.3002e+06
	0.105   400 30  5.2684e+06
	0.205   400 30  5.5534e+06
	0.305   400 30  6.1292e+06
	0.405   400 30  6.6702e+06
	0.505   400 30  7.2192e+06
	0.605   400 30  6.9306e+06
	0.705   400 30  5.5206e+06
	0.805   400 30  4.0857e+06
	0.905   400 29  5.1603e+06
	
	0.005   450 30  8.0286e+06
	0.105   450 30  5.8948e+06
	0.205   450 30  5.9761e+06
	0.305   450 30  6.5749e+06
	0.405   450 30  7.6008e+06
	0.505   450 30  8.2071e+06
	0.605   450 30  7.7313e+06
	0.705   450 30  8.0063e+06
	0.805   450 29  8.7191e+06
	0.905   450 28  8.9783e+06
};
\end{axis}    

\draw [-,  very thick , fill = white] ($(nnOUT)+(\rectSize,0)$) -- ++ (1.6,0) node [above, text width=1cm, align=center,  midway] { \textit{Inverse \\ GFT}};

\foreach \i in {1,...,4}
\draw [->] (input-\i) -- (hidden-\i);
\foreach \i in {1,...,4}
\foreach \j in {1,...,3}
\draw [->] (hidden-\i) -- (output-\j);

\coordinate (C) at ($(x0)+(13,0)$);

\draw  [->,  very thick , fill = white] (C.east) -- ++ (0.5,0) node [right, text width=0.2cm] {${\mathbf{F}_\textrm{pred}}$};
\end{tikzpicture}

%% file: tikz/spectralfiltering.tex
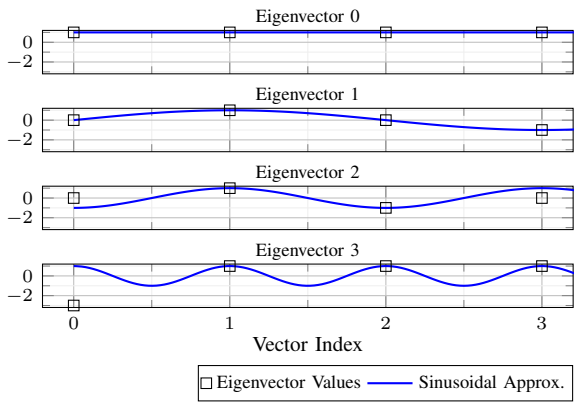
\begin{figure}[t!]
\def \dist {0.12}
\def \hscale {0.25}
\vspace{2pt}
\begin{tikzpicture}
\begin{axis}[
xmin = -0.2, xmax = 3.2,
ymin = -3.2, ymax = 1.2,
y label style={at={(axis description cs: 0.1,.5)}, anchor=south},
xtick distance = 1, 
xticklabels={,,},
ytick distance = 2,
tick label style={font=\scriptsize} ,
title style={yshift=-8pt, font=\scriptsize},
title = Eigenvector 0,
grid = both,
minor tick num = 1,
major grid style = {lightgray},
minor grid style = {lightgray!25},
width = \columnwidth,
height = \hscale\columnwidth]
\addplot[
domain = 0:5,
samples = 100,
smooth,
thick,
blue,
] {1};
\addplot [only marks, mark= square] table {
	0 1
	1 1
	2 1
	3 1
};
\end{axis}
\begin{axis}[
yshift=-\dist\columnwidth,
xmin = -0.2, xmax = 3.2,
ymin = -3.2, ymax = 1.2,
y label style={at={(axis description cs: 0.1,.5)}, anchor=south},
tick label style={font=\scriptsize} ,
xtick distance = 1,
xticklabels={,,},
ytick distance = 2,
title style={yshift=-8pt, font=\scriptsize},
title = Eigenvector 1,
grid = both,
minor tick num = 1,
major grid style = {lightgray},
minor grid style = {lightgray!25},
width = \columnwidth,
height = \hscale\columnwidth]
\addplot[
domain = 0:5,
samples = 100,
smooth,
thick,
blue,
] {sin(deg(0.5*x*pi)))};
\addplot [only marks, mark= square] table {
	0  0
	1  1
	2  0
	3  -1
};
\end{axis}
\begin{axis}[
yshift=-2*\dist\columnwidth,
xmin = -0.2, xmax = 3.2,
ymin = -3.2, ymax = 1.2,
y label style={at={(axis description cs: 0.1,.5)}, anchor=south},
tick label style={font=\scriptsize} ,
title style={yshift=-8pt, font=\scriptsize},
title = Eigenvector 2,
xtick distance = 1,
xticklabels={,,},
ytick distance = 2,
grid = both,
minor tick num = 1,
major grid style = {lightgray},
minor grid style = {lightgray!25},
width = \columnwidth,
height = \hscale\columnwidth]
\addplot[
domain = 0:5,
samples = 200,
smooth,
thick,
blue,
] {sin(deg((x-0.5)*pi)))};
\addplot [only marks, mark= square] table {
	0  0
	1  1
	2  -1
	3  0
};
\end{axis}
\begin{axis}[
legend style={at={(1,0)}, xshift=-0.0cm,yshift =-1.25cm,
	anchor=south east, nodes=right, minimum size=0.4cm, inner sep=1pt},
legend columns=2,
legend entries={{\scriptsize Eigenvector Values},{\scriptsize Sinusoidal Approx.}},
yshift=-3*\dist\columnwidth,
xmin = -0.2, xmax = 3.2,
ymin = -3.2, ymax = 1.2,
y label style={at={(axis description cs: 0.1,.5)}, anchor=south},
x label style={at={(axis description cs: 0.5,-.3)}, anchor=south},
tick label style={font=\scriptsize} ,
title style={yshift=-8pt, font=\scriptsize},
title = Eigenvector 3,
xlabel = \footnotesize  Vector Index,
xtick distance = 1,
ytick distance = 2,
grid = both,
minor tick num = 1,
major grid style = {lightgray},
minor grid style = {lightgray!25},
width = \columnwidth,
height = \hscale\columnwidth]
\addplot [only marks, mark= square] table {
	0  -3
	1  1
	2  1
	3  1
};
\addplot[
domain = 0:5,
samples = 100,
smooth,
thick,
blue,
] {cos(deg((x)*2*pi)))};
\end{axis}
\end{tikzpicture}
\caption{Graph Laplacian eigenvectors resulting from the eigendecomposition of a four-node star graph. The eigenvectors are used as basis functions in the GFT. The eigenvectors are approximated by best-fit sinusoidal functions, which form the basis in the sequential Fourier transform.}
\label{fig:sinusoidalApprox}
\vspace{-18pt}
\end{figure}